\documentclass{article}

\usepackage[margin=1in]{geometry}
\usepackage[slantedGreek]{mathpazo}
\usepackage[round]{natbib}
\usepackage{graphicx,amsmath,amsfonts,dsfont}
\usepackage{float}
\usepackage{xskak}
\usepackage{chessboard}
\usepackage{tikz}

\title{On The Value of Chess Squares}
\author{	
	\makebox[.4\linewidth]{Aditya Gupta}\\\textit{\small  Chess ED}\\\textit{\small  Chicago, IL}\\\and
	\makebox[.4\linewidth]{Shiva Maharaj}\\\textit{\small  Chess ED}\\\textit{\small  Chicago, IL}\\\and
	\makebox[.4\linewidth]{Nicholas Polson\footnote{Email: ngp@chicagobooth.edu}}\\\textit{\small  Booth School of Business}\\
	\textit{\small  University of Chicago}\\
	\and 
	\makebox[.4\linewidth]{Vadim Sokolov}\\
	\textit{\small  Department of Systems Engineering }\\
	\textit{\small  and Operations Research}\\
	\textit{\small  George Mason University}\\
}

\graphicspath{{fig/}}

\begin{document}

\maketitle
\emph{Chess is not a game. Chess is a well-defined form of computation. You may not be able to work out the answers, but in theory, there must be
  a solution, a right procedure in any position.    \hspace{1.0in} ---John von Neumann}
\begin{abstract}
  \noindent  We propose a neural network-based approach to calculate the value of a chess square-piece combination. Our model takes a triplet (Color, Piece, Square) as an input and calculates a value that measures  the advantage/disadvantage of having this piece on this square. Our methods build on recent advances in chess AI, and can accurately assess the worth of positions in a game of chess. The conventional approach assigns fixed values to pieces $(\symking=\infty, \symqueen=9, \symrook=5, \symbishop=3, \symknight=3, \sympawn=1)$. We enhance this analysis by introducing marginal valuations. We use deep Q-learning to estimate the parameters of our model. We demonstrate our method by examining the positioning of Knights and Bishops, and also provide valuable insights into the valuation of pawns. Finally, we conclude by suggesting potential avenues for future research.
\end{abstract}

\vspace{0.1in}

\vspace{0.1in}

\section{Introduction}\label{sec:intro}

The development of Chess AI can be traced back to pioneering work by \cite{turing45}, \cite{Shannon50}, \cite{Bronowski_Bragg_Gower_2012a}, \cite{botvinnik2012computers}, and \cite{good1988five}. While Shannon's approach was primarily based on trial and error, leading to the ``learning" of the optimal policy, Turing and Champernowne emphasized marginal piece valuation. Notably, the Shannon number estimates the number of possible chess board states to be $10^{152}$, out of which $10^{43}$ are legal states. Turing and Champernowne's positional evaluation functions included criteria such as piece mobility, piece safety, king mobility, king safety, and the ability to castle. Good took on a more linear-programming based approach to value the chess pieces. 

All of these methods have now been replaced with either $\alpha-\beta$ search (\cite{romstad2011stockfish}) or deep learning \citep{dalgaard2020global, pascutto2022leela}. Deep learning models allow to  evaluate state-dependent objective functions through learning. A reinforcement learning algorithm is used to estimate the parameters of deep learning models using trial-and-error self-play approach  \citep{silver2016mastering,maharaj2022chess}.   A testament to this advancement is AlphaZero, which, through self-play, can deduce effective solution paths and ``learn" chess in under four hours without any prior knowledge (\cite{dean2012large} and \cite{silver2017mastering}).

While much recent work has been done in Chess AI \citep{kapicioglu2020chess2vec, maharaj2021karpov, maharaj2022chess}, the question of the value of a chess square has not yet been explored. The goal of this study is to use neural networks to measure the advantages/disadvantages, particularly the $Q-$value of state $s \in \text{{Color}} \times \text{{Piece}} \times \text{{Square}}$, offered by the state combination of particular chess squares with different pieces. We argue that from the perspective of coaching and training a chess player, it is more useful to understand the value of squares rather than knowing what move needs to be made in any given position. 

\begin{figure}[H]
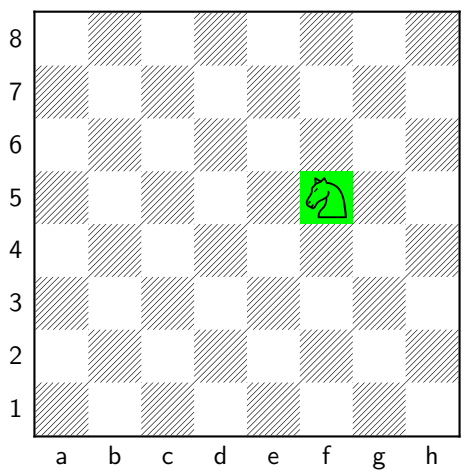

  \center
  \setchessboard{
    showmover=false,
    moverstyle=triangle,
    pgfstyle=color,
    pgfcolor=blue,
    markfield={},
    color=green,
    backfields={f5},
  }
  \newgame
  \fenboard{8/8/8/5N2/8/8/8/8 w - - 0 1}

  \chessboard
  \caption{Each square on the chess board can be defined by the letter of its column and the number of its row. For example, the white knight on the highlighted square in this diagram is located at $f5$}
  \label{fig:chess_notation}
\end{figure}

For example, the notion that certain state combinations, such as having a White $ \symknight $ on $f5$ (as can be seen in Figure \ref{fig:chess_notation})provide an advantage to White players is a widely held belief in the world of chess.  We analyze these key combinations to see whether the games of high-level chess grandmasters provide merit to this belief. Our investigation will shed light on the strategic nuances and patterns that emerge from such positions and contribute to the understanding of chess at the highest level of play.

To value pieces on squares, we create a Neural Network to analyze a dataset of Grandmaster games and make predictions regarding winning probabilities. This approach uses Centipawn evaluations for specific subsets of chess states involving Knight and Bishop pieces. The centipawn is used in chess as the unit of measure for the advantage offered to one side, with one centipawn being equal to roughly a hundredth of a pawn.  The results show that our model successfully generated predictions for Knights and Bishops. The predictions provided valuable insights into the advantages and disadvantages associated with different states and positions on the chessboard. In addition, many pieces of preexisting knowledge based on anecdotal evidence were confirmed by our analysis. For example, the analysis revealed that Knights placed in the corners of the board had lower winning probabilities, likely due to their limited mobility and restricted influence. On the other hand, as Knights moved closer to the opponent's side, their positional value tended to increase, potentially allowing them to infiltrate enemy territory and exert greater control over the game. The study's results enhance the understanding and confirmation of chess strategies and gameplay dynamics, aiding in strategic decision-making and the evaluation of different gameplay approaches.

Several chess maxims, known to be true from the anecdotal experience of chess players, are reflected in our neural network predictions. For example, Pawns are observed to gain in value as they cross the 4th rank, highlighting the significance of advancing pawns beyond this milestone. Pawns positioned on the 5th rank are particularly powerful, contributing to central control and potential attacking opportunities. Pawns on the 6th rank, especially when supported by a pawn on the 5th rank, become highly threatening. Edge pawns tend to be weaker compared to central pawns, emphasizing the importance of controlling central squares. 

We uncovered several chess concepts previously not recognized as common knowledge. Notably, we observed that advanced kingside pawns pose a greater threat than their queenside counterparts. Such uncommon ideas add to the arsenal of human knowledge and allow chess players to better understand and gain strategic insights into their chess play.

Important squares for the white pawn are identified by examining the highest Centipawn evaluation $c(s)$ values in each column. The squares $e4$, $h4$, $c5$, and $h6$ are highlighted as critical positions for white pawns. Occupying these squares provides advantages, such as central control, support for piece development, and potential attacking opportunities.

Similarly, for black pawns, the squares $f5$, $d5$, $c4$, $d3$, and $f3$ emerge as key positions. Placing pawns on these squares enhances black's control of central areas, supports piece coordination, and enables counter-play against white's position.

Understanding the significance of these key squares and applying the derived insights allows players to make informed decisions regarding pawn placement, pawn breaks, and strategic plans. This knowledge empowers players to optimize their pawn structures, control critical areas of the board, and leverage their pawns to gain a competitive advantage in the game. It also confirms much of previously anecdotal-based knowledge in a more computational and rigorous method.

The rest of the paper is outlined as follows. Section \ref{sec:prev-work} provides connections with previous literature. Section \ref{sec:methods} goes over the methods we used. Section \ref{sec:applications} provides an application of the proposed methods to Grandmasters and Magnus Carlsen (World Chess Champion, 2013-2023). Section \ref{sec:pawns} provides an application to Pawns. Finally, Section \ref{sec:conclusion} concludes.
\subsection{Connections with Previous Work}\label{sec:prev-work}
In the field of Chess AI, previous research has primarily focused on predicting the probabilities of winning $w(s)$ and Centipawn evaluations $c(s)$ for more simplified states\citep{maesumi2020playing}. \cite{gupta2023determining} explored simpler states where $s$ belongs to the set of $s \in \text{Piece}$, with no consideration given to the position of a chess piece. In their work, they utilized Logistic Regression methods to determine the value of a chess piece by creating a model that predicts the outcome of a game based on existing piece imbalances in a given position. A recent lichess study also tried similar approaches \cite{ubdip2022finding}, \cite{ubdip2022comments}.

Building upon this previous work, our research extends the scope by proposing an augmented state representation $s$ that encompasses $\text{{Color}} \times \text{{Piece}} \times \text{{Square}}$, thereby incorporating the square (location) information as an additional component of the state. This augmentation enables a more comprehensive understanding of the game dynamics by considering both the piece and its position on the board. Furthermore, we employ Neural Networks as our chosen methodology, allowing us to capture and model the intricate relationships between the state $s$ and its corresponding Centipawn evaluation $c(s)$.

One crucial distinction between our proposed approach and previous methodologies lies in the predictive target. While prior research focused on predicting the binary outcome of the game (win or loss), our proposed model aims to predict the Centipawn evaluation $c(s)$ instead. By doing so, we shift the focus towards assessing the advantage or disadvantage of a particular chess position, providing more granular information beyond a simple win/loss prediction.

By using the augmented state representation and employing Neural Networks, our proposed model offers a more comprehensive and nuanced analysis of the chess game. This allows us to capture the intricate interplay between the color, piece type, square, and Centipawn evaluation, providing a deeper understanding of the factors influencing the game's outcome.

This paper extends the work of \cite{maharaj2022chess} and proposes novel architectures that can predict the probabilities of winning $w(s)$ and Centipawn evaluations $c(s)$ for all possible states $s \in \text{{Color}} \times \text{{Piece}} \times \text{{Square}}$. While previous work focused on specific subsets of states, particularly those related to gambits, our approach seeks to encompass the entire chessboard by incorporating the color, piece type, and square information into a comprehensive state representation.

By embracing a wider scope of analysis that covers all possible states, our research aims to provide a more comprehensive understanding of the game, surpassing the limitations imposed by narrow subsets. To achieve this, we employ advanced techniques, such as Neural Networks, to capture the intricate relationships between the components of a state and the corresponding probabilities of winning $w(s)$ and Centipawn evaluations $c(s)$. This allows us to offer valuable insights into the dynamics of chess gameplay across a vast array of states, thereby providing a more holistic and comprehensive analysis.

Through our research, we strive to advance the field by developing robust and effective models capable of accurately predicting the probabilities of winning and assessing the Centipawn evaluations for any given state. By considering the full spectrum of states represented by $\text{{Color}} \times \text{{Piece}} \times \text{{Square}}$, our proposed architectures pave the way for a deeper understanding of chess strategies. They enable us to evaluate the efficacy of these strategies and unravel the intricacies of the game, ultimately contributing to the development of more sophisticated and intelligent Chess AI systems.

\section{Value of Chess Squares and Pieces}\label{sec:methods}

Traditional chess literature often quotes the values of pieces as:
\[
  ( \symking, \symqueen , \symrook, \symbishop , \symknight , \sympawn ) = ( \infty , 9 , 5 , 3, 3 ,1 ).
\]
However, modifications to these conventional valuations have been proposed. For instance, \cite{gupta2023determining} adjusted these values using Machine Learning techniques, resulting in:
\[
  ( \symking, \symqueen , \symrook, \symbishop , \symknight , \sympawn ) = ( \infty , 8.9 , 4.6 , 3.3, 3 ,1 ).
\]
Similarly, a recent study conducted on Lichess \citep{ubdip2022finding} refined these values further to:
\[
  ( \symking, \symqueen , \symrook, \symbishop , \symknight , \sympawn ) = ( \infty , 9.82 , 4.93 , 3.28, 3.16 , 1 ).
\]
Building upon these studies, our research aims to enhance the valuation system by incorporating the square position into the state vector.

Our work will provide values for states consisting of a combination of pieces and squares. For example, we may wish to assess the value of a fianchettoed bishop on the queen's side. The proposed method measures the advantage/disadvantage for states of the form  $s \in \text{{Color}} \times \text{{Piece}} \times \text{{Square}}$. For a Bishop on $b2$, we denote this value by  $$ V ( \symbishop, b2 ) $$ while a white knight on an outpost such  as $f5$ is denoted by $ V (  \symknight , f5) $. The valuation of pieces plays a pivotal role in determining strategic decisions. The value attributed to a piece, such as the knight, is intrinsically tied to the probability of winning, as assessed by a chess engine. We can express this relationship through the identity:
\[
  V ( \symknight  ) = \sum_{ \text{position} }  V ( \symknight , \text{position} ),
\]
where the summation is executed over all plausible future positions. This elucidates that the initial valuation of the knight, often referred to as \( V ( \symknight  ) = 3 \), is an aggregate of its utility across the entire game. As the game progresses and pieces are maneuvered, these values experience marginal shifts. Our primary objective is to accurately determine position-specific values, such as \( V (  \symknight , f5) \).

A brief overview of the methods in this paper are as follows:

\begin{enumerate}
    \item A comprehensive dataset \( \mathcal{D} \) is systematically created, encompassing a diverse range of chessboard states \( s \). This dataset is derived from an extensive chess database, as described in section \ref{sec:data-preparation}.
    \item Utilizing the state-of-the-art chess engine, Stockfish 16 \cite{romstad2022stockfish}, an evaluation is derived for each state \( s \) in \( \mathcal{D} \). This evaluation yields the true value \( c(s) \), indicative of the strategic strength of the corresponding chessboard state.
    \item A neural network architecture is designed and implemented. This network is specifically tailored to approximate the function \( V: s \mapsto c(s) \), where \( V(s) \) represents the predicted value of state \( s \) as determined by the neural network.
    \item In the applications section, the constructed neural network is rigorously tested by employing it to predict \( V(s) \) for a myriad of states \( s \), thereby assessing its results in relation to anecdotal chess maxims.
\end{enumerate}

\subsection{Centipawn Evaluation and Optimal Play}
In our approach, we begin by formalizing the theoretical functions used in $Q$-learning. The value function, denoted as $V(s)$, represents the probability of winning the game given a specific state $s$. This state $s$ belongs to the set $\text{{Color}} \times \text{{Piece}} \times \text{{Square}}$, and it is worth emphasizing that $V(s)$ is calculated with respect to the color parameter in any given state.

To assess any legal chess position, we derive a Centipawn evaluation denoted as $c(s)$. The Centipawn serves as a measurement unit for evaluating the advantage in chess, where one Centipawn is equal to 1/100 of a pawn. The win probability $w(s)$ can be directly obtained from $c(s)$ by using the following equation:

\[
  w(s) = \mathbb{P}(\text{{winning}}|s) = \frac{1}{1+10^{-c(s)/4}}, \quad \text{{and}} \quad c(s) = 4\log_{10}\left(\frac{w(s)}{1-w(s)}\right).
\]

For example, if White has a $ c(s) =0.2 $ advantage, then the win probability is $w(s) = 0.526$.

To address the sequential decision problem, we employ the dynamic programming technique known as Q-learning(\cite{watkins1992q}, \cite{o2018uncertainty}, \cite{clifton2020q}). This methodology involves breaking down the decision problem into smaller sub-problems. A key principle utilized in Q-learning is Bellman's principle of optimality, which states:

\vspace{0.1in}

\begin{center}
  \textit{Bellman Principle of Optimality: An optimal policy has the property that whatever the initial state and initial decision are, the remaining decisions must constitute an optimal policy with regard to the state resulting from the first decision. (Bellman, 1957)}
\end{center}

\vspace{0.1in}

To solve this sequential decision problem, we employ Backwards Induction (\cite{priest2000logic}), which determines the most optimal action at the last node in the decision tree (i.e., the checkmate position). Utilizing this information, we can then determine the best action for the second-to-last decision point, and this process continues backward until we identify the optimal action for every possible situation, effectively solving the Bellman equation.

In recent years, the field of artificial intelligence has witnessed significant advancements, particularly in the realm of AI algorithms like deep learning(\cite{lecun2015deep}, \cite{yan2015deep}, \cite{saadat2020advancements}, \cite{vargas2017deep}), leading to advances in other fields (\cite{gupta2023forecast, gupta2023analysis,gupta2023using}) alongside the development of remarkably powerful computer chess engines. These technological breakthroughs have revolutionized the way we evaluate and understand chess positions, enabling us to delve into the intricacies of the game with unparalleled precision.

One notable achievement stemming from these advancements is the ability to accurately assess chess positions. By leveraging AI algorithms, particularly deep learning techniques, we can now analyze and comprehend chess moves and strategies in a manner that was previously unimaginable. These algorithms have been specifically designed to process vast amounts of data, learn from patterns, and make informed decisions, ultimately resulting in highly accurate evaluations of chess positions.

Moreover, the advent of advanced computer chess engines, exemplified by the likes of Stockfish 15 \cite{romstad2011stockfish}, has played a pivotal role in shaping the landscape of chess analysis and study. These engines, meticulously crafted through a combination of cutting-edge algorithms and extensive programming, have transformed the way chess is played and understood.

The emergence of chess engines has effectively shifted the burden from human players and theorists to these intelligent systems. By leveraging their computational power and algorithmic prowess, chess engines have assumed the responsibility of assessing various lines of play, thus solving the Bellman equation.

By adhering to Bellman's optimality condition, computer chess engines fulfill the requirements of possessing complete knowledge about the chess environment and evaluating all possible actions and their consequences. Through this rigorous analysis, they provide insights into the optimal move in a given position.

\subsection{Q-Values}

For a given $c(s)$ values, the corresponding $Q$-value represents the probability of winning, given a move $a$ in a given state $s$, by following the optimal Bellman path thereafter:

\[Q(s, a) = \mathbb{P}(\text{{winning}}|s, a).\]

To address the optimal sequential decision problem, we employ $Q$-learning, which calculates the $Q$-matrix (\cite{korsos2014analyzing}, \cite{polson2015bellman}), denoted as $Q(s, a)$ for a given state $s$ and action $a$. The $Q$-value matrix describes the value of performing action $a$ and then acting optimally thereafter. The current optimal policy and value function can be expressed as follows:

\[V(s) = \underset{a}{\max} \, Q(s, a) = Q(s, a^*(s)) \; \; {\rm where} \; \; a^*(s) = \text{{argmax}}_a \, Q(s, a).\]

The policy function establishes the optimal mapping from states to actions, and by substituting the $Q$-values, we obtain the value function for a given state.

In Section \ref{sec:neural-network}, we introduce a Neural Network architecture designed specifically for predicting the value of $c(s)$ given the state $s$. By harnessing the predictive capability of this Neural Network, we can subsequently determine the probability of a player winning, denoted as $w(s)$, based on their corresponding state $s$.

The Neural Network model comprises interconnected layers, including an input layer that accepts the state $s$ as input. Through a series of computations within the hidden layers, the model captures complex relationships and patterns inherent in the input data. Ultimately, the output layer produces the predicted value of $c(s)$.

By employing this trained Neural Network model, we can make predictions of $c(s)$ for unseen states $s$. These predicted values can then be utilized to compute the probability of a player winning, denoted as $w(s)$. The specific relationship between $c(s)$ and $w(s)$ is contingent upon the characteristics and dynamics of the chess game under analysis.

With the ability to predict $w(s)$, we gain valuable insights into the probability of a player winning based on their current state $s$. This information can be harnessed in various ways, including evaluating strategic moves, assessing the overall advantage or disadvantage of specific board configurations, and guiding decision-making during gameplay.

The Neural Network's capacity to capture intricate patterns and relationships within the input data significantly contributes to more accurate predictions and a deeper understanding of the dynamics of the chess game. By incorporating the predicted values of $c(s)$ and computing the corresponding probabilities of winning, we enhance our analytical capabilities and facilitate informed decision-making in the context of chess gameplay.

\subsection{Neural Network Architecture}\label{sec:neural-network}
We design a 3-layer Neural Network aimed at determining the $Q$-values, by predicting the value of a chess square and piece combination, denoted as $c(s)$ for $s \in \text{{Color}} \times \text{{Piece}} \times \text{{Square}}$. This model incorporates a hyperbolic tangent ($\tanh$) activation function as a key component of its architecture. The use of the $\tanh$ function, with its symmetric nature around the origin, ensures a balanced output range, making it apt for this application.

While the model can be made more sophisticated, the simplicity of this 3-layer architecture was chosen deliberately. The primary focus is on the novel method of determining values for chess squares and pieces, rather than leveraging computational prowess. By keeping the model straightforward, we ensure that results are more interpretable and that the emphasis remains on the innovation behind our method.


To ensure effective training of the model, we curate a meticulously crafted dataset. This dataset consists of two essential elements: the state information, represented by $s$, and the corresponding centipawn loss (CPL) recorded for each state. Centipawn loss refers to the average difference in evaluation points between a chess engine's top move and the move played by a player, expressed in hundredths of a pawn. Lower centipawn loss values indicate more accurate play, while higher values suggest greater inaccuracies or mistakes. The state information encompasses the variables discussed earlier that define the given chessboard position. 

Through supervised learning using this dataset, the model learns to associate the given state information with the corresponding CPL. Consequently, it acquires the ability to predict the CPL based on the provided state information as input. This training process involves iteratively adjusting the model's parameters to minimize the disparity between its predictions and the actual CPL values present in the training dataset.

The neural network processes an input vector \( \mathbf{x} \) using a series of transformations:

\[
f_1(\mathbf{x}) = \tanh(\mathbf{W}_1 \mathbf{x} + \mathbf{b}_1)
\]

Where \( \mathbf{W}_1 \) is the weight matrix of the first layer with 64 neurons, and \( \mathbf{b}_1 \) is its bias vector. The activation function used is the hyperbolic tangent function, \( \tanh \). The output of this layer is \( \mathbf{a}_1 = f_1(\mathbf{x}) \).

\[f_2(\mathbf{a}_1) = \tanh(\mathbf{W}_2 \mathbf{a}_1 + \mathbf{b}_2)\]

For the second layer, \( \mathbf{W}_2 \) represents the weight matrix of the 32 neurons, and \( \mathbf{b}_2 \) is its bias vector. The activation function remains \( \tanh \). The output of this layer is \( \mathbf{a}_2 = f_2(\mathbf{a}_1) \).

\[f_3(\mathbf{a}_2) = \mathbf{W}_3 \mathbf{a}_2 + \mathbf{b}_3\]

The third and final layer has only one neuron, and it directly outputs the weighted sum without any activation function. Therefore, \( f_3(\mathbf{a}_2) \) gives the network's output for a given input \( \mathbf{x} \).

Thus, the entire network can be represented by the composite function:

\[
f(\mathbf{x}) = f_3(f_2(f_1(\mathbf{x})))
\]

Where \( \mathbf{W}_i \) are weight matrices, \( \mathbf{b}_i \) are bias vectors for each respective layer.

We use an Adam optimizer with a MSE loss. The model processes the input through the layers, applying the weighted sums and hyperbolic tangent activations sequentially, to finally produce the desired output value.

\subsection{Data}\label{sec:data-preparation}

In order to train the Neural Network effectively, a training dataset is constructed, comprising two essential components. This dataset consists of elements that contain both the state information denoted by $s$, as well as the corresponding evaluation associated with that particular state.

To gather the necessary chess game data for analysis, a vast mega database containing millions of previously played chess games is utilized. Within this database, each game is represented using the Portable Game Notation (PGN) notation, which allows for standardized representation and compatibility with various chess software and applications.

The process of constructing the training dataset involves parsing and evaluating all positions $p$ within each game. The Forsyth-Edwards Notation (FEN) is employed to determine the location of relevant chess pieces within each position $p$. As a result, all states $s \in p$ are extracted and added to the training dataset. To navigate through the moves of each chess game systematically, the Python Chess library is utilized. This library provides a comprehensive set of functions and classes specifically designed for working with chess games and positions, enabling efficient traversal of the stored games in the database.

For every position $p$ within the dataset, an evaluation is obtained. To accomplish this, the research incorporates the Stockfish engine, a widely recognized and powerful chess engine. Stockfish employs advanced algorithms and evaluation functions to assess the strength of positions. By leveraging the capabilities of Stockfish, the training dataset can determine the evaluation of each position $p$ on the chessboard accurately.

Finally, this evaluation is associated with all states $s \in p$, resulting in a comprehensive dataset that encompasses both the state $s$ and the evaluation associated with the position $p$ from which $s$ was derived. This dataset serves as the foundation for training the Neural Network, enabling it to learn and make informed decisions based on the provided state information.

\section{Piece Evaluation: Knights \& Bishops}\label{sec:applications}

In this study, our proposed model is applied to a comprehensive dataset comprising over 2000 Grandmaster games. The primary objective is to predict the probabilities of winning $w(s)$ and Centipawn evaluations $c(s)$ for a specific subset of states, namely those denoted by $\{ (c, p, sq) \in s : p \in \{\text{{Knight}}, \text{{Bishop}}\} \}$. Although our focus is initially on the Knight and Bishop pieces, it is important to note that the model can be expanded to encompass all pieces, offering a broader analysis of the game.

To provide a visual representation of the predicted values, heat maps are generated for both $w(s)$ and $c(s)$ corresponding to each valid combination within the specified subset. These heat maps offer a comprehensive overview of the probabilities of winning and Centipawn evaluations associated with the Knight and Bishop pieces in different states.

\begin{figure}[H]
  \centering
  \begin{minipage}[b]{0.45\linewidth}
    \centering
    \includegraphics[width=\textwidth]{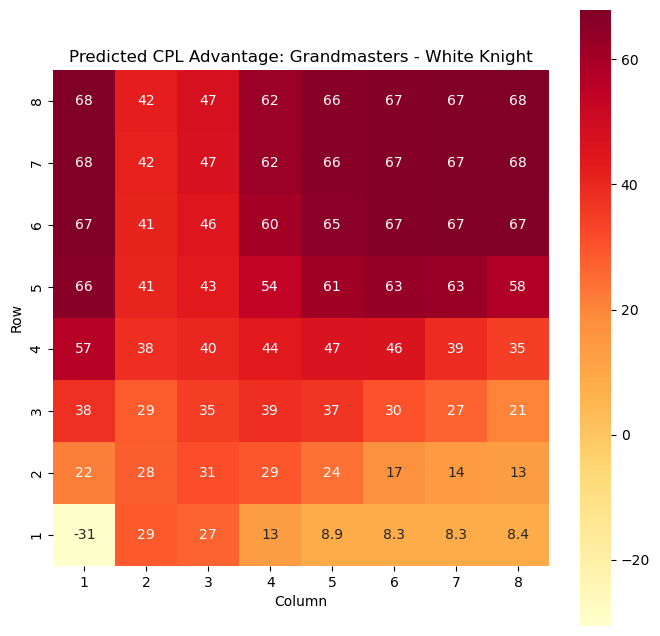}
    \caption{Predicted CPL Advantage $c(s)$ Offered by White Knight in Grandmaster Games}
    \label{fig:gm-wk-1}
  \end{minipage}
  \hspace{0.5cm}
  \begin{minipage}[b]{0.45\linewidth}
    \centering
    \includegraphics[width=\textwidth]{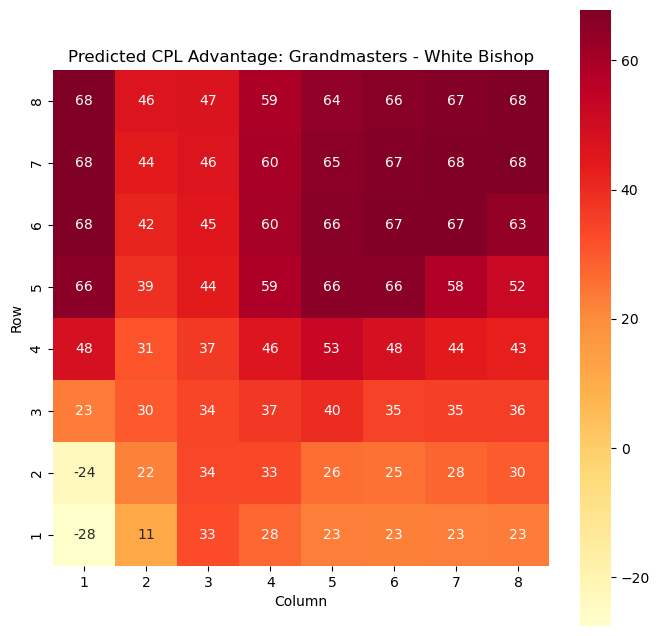}
    \caption{Predicted CPL Advantage $c(s)$ Offered by White Bishop in Grandmaster Games}
    \label{fig:gm-wb-1}
  \end{minipage}
\end{figure}

\begin{figure}[H]
  \centering
  \begin{minipage}[b]{0.45\linewidth}
    \centering
    \includegraphics[width=\textwidth]{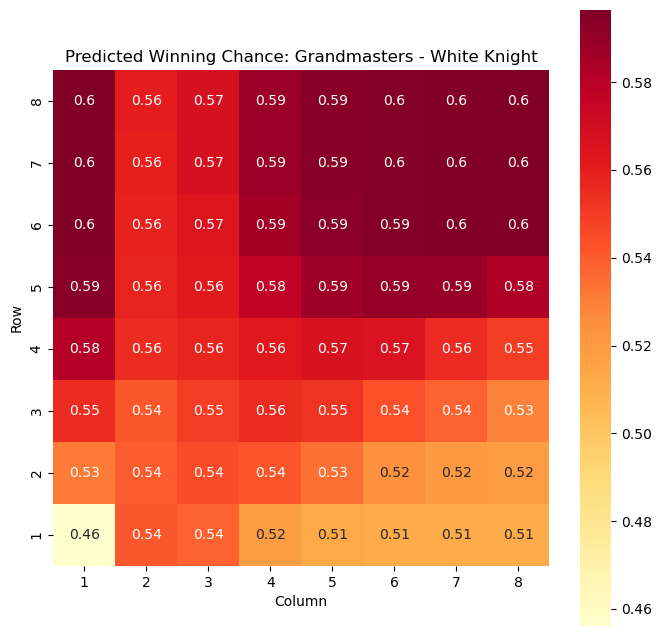}
    \caption{Predicted Winning Chance $w(s)$ Offered by White Knight in Grandmaster Games}
    \label{fig:gm-wk-2}
  \end{minipage}
  \hspace{0.5cm}
  \begin{minipage}[b]{0.45\linewidth}
    \centering
    \includegraphics[width=\textwidth]{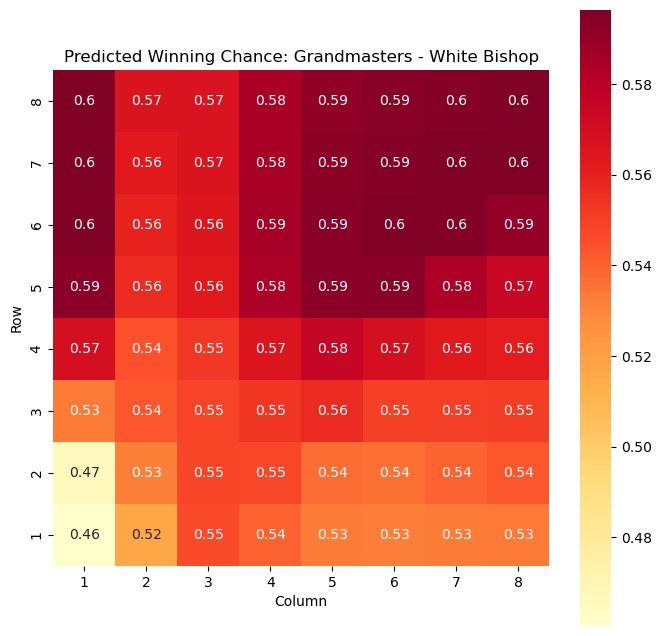}
    \caption{Predicted Winning Chance $w(s)$ Offered by White Bishop in Grandmaster Games}
    \label{fig:gm-wb-2}
  \end{minipage}
\end{figure}

To illustrate the efficacy of our model, we first employ it to predict the Centipawn evaluations $c(s)$ specifically for states where the color $c$ is White and the piece $p$ is Knight or Bishop. The resulting predictions are showcased in Figure \ref{fig:gm-wk-1} and Figure \ref{fig:gm-wb-1}, providing valuable insights into the relative advantages or disadvantages of such states. Building upon this, we further use $c(s)$ to derive the corresponding probabilities of winning $w(s)$ for these specific states. The model-generated probabilities are visualized in Figure \ref{fig:gm-wk-2} and Figure \ref{fig:gm-wb-2}, offering a clear representation of the likelihood of White winning the game given the occurrence of the specified state $s$.

By leveraging our proposed model, we gain a deeper understanding of the dynamics of the game, specifically in relation to the Knight and Bishop pieces within the context of the White color. This analysis not only facilitates strategic decision-making but also provides a basis for evaluating the effectiveness of various gameplay approaches. Moreover, the model's expandability to encompass all pieces allows for a comprehensive examination of the game across different states, enabling us to uncover additional insights and enhance the overall understanding of chess strategies and pawn skeletons, as Nimzowitch had surmised in his seminal work (\cite{nimzowitsch2022my}).

\begin{figure}[H]
  \centering
  \begin{minipage}[b]{0.45\linewidth}
    \centering
    \includegraphics[width=\textwidth]{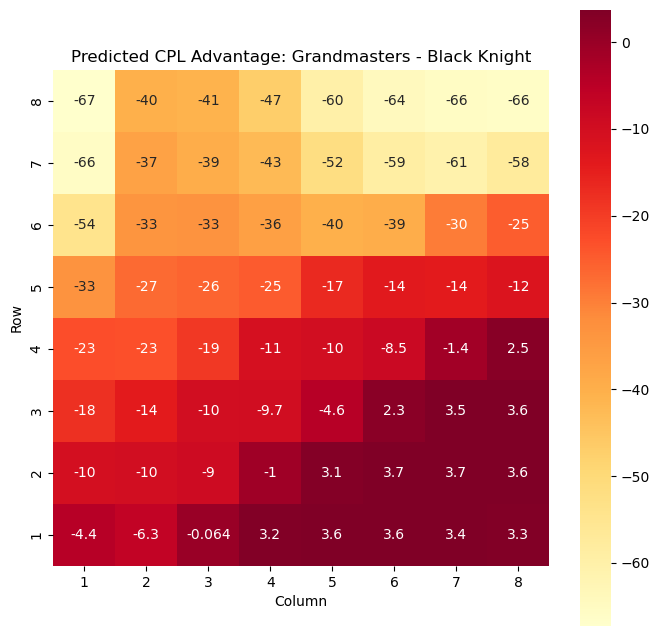}
    \caption{Predicted CPL Advantage $c(s)$ Offered by Black Knight in Grandmaster Games}
    \label{fig:gm-bk-1}
  \end{minipage}
  \hspace{0.5cm}
  \begin{minipage}[b]{0.45\linewidth}
    \centering
    \includegraphics[width=\textwidth]{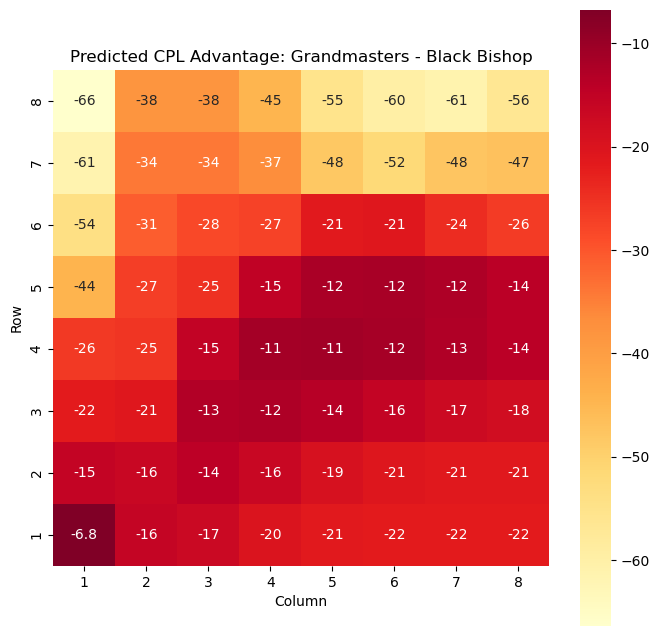}
    \caption{Predicted CPL Advantage $c(s)$ Offered by Black Bishop in Grandmaster Games}
    \label{fig:gm-bb-1}
  \end{minipage}
\end{figure}

\begin{figure}[H]
  \centering
  \begin{minipage}[b]{0.45\linewidth}
    \centering
    \includegraphics[width=\textwidth]{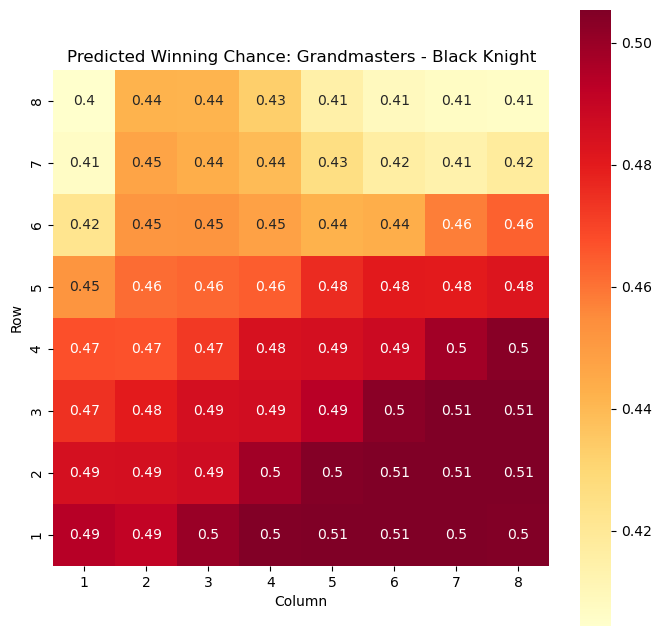}
    \caption{Predicted Winning Chance $w(s)$ Offered by Black Knight in Grandmaster Games}
    \label{fig:gm-bk-2}
  \end{minipage}
  \hspace{0.5cm}
  \begin{minipage}[b]{0.45\linewidth}
    \centering
    \includegraphics[width=\textwidth]{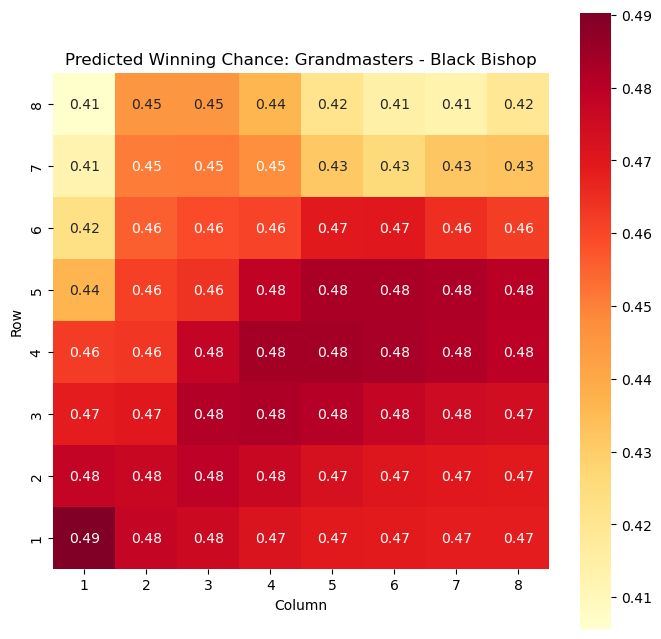}
    \caption{Predicted Winning Chance $w(s)$ Offered by Black Bishop in Grandmaster Games}
    \label{fig:gm-bb-2}
  \end{minipage}
\end{figure}

The model is then used to determine $c(s)$ and $w(s)$ for states $\{ (c, p, sq) \in s : c = \text{{"Black"}}, p = \text{{"Knight", "Bishop"}} \}$, as can be seen in Figure \ref{fig:gm-bk-1}, Figure \ref{fig:gm-bk-2}, Figure \ref{fig:gm-bb-1}, and Figure \ref{fig:gm-bb-2} respectively.

Key squares for the Bishops can be seen in Figure \ref{fig:key-bish}:
\begin{figure}[htb]
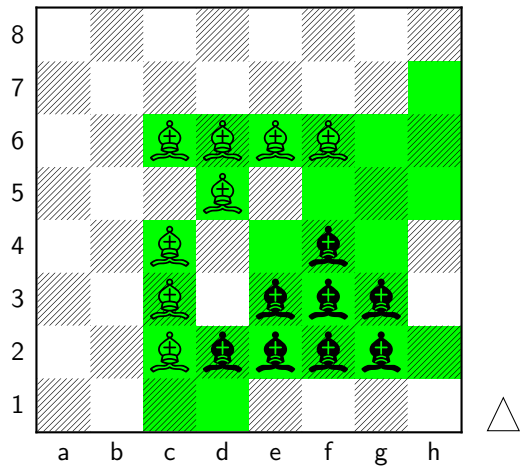

  \center
  \setchessboard{
    showmover=true,
    moverstyle=triangle,
    pgfstyle=color,
    pgfcolor=blue,
    markfield={},
    color=green,
    backfields={d2,e2,f2,g2,e3,f3,g3,f4,g5,h6,h2,g4,h5,d1,c1, c6,d5,e4,f5,g6,h7,c2,c3,c4,e6,d6,f6},
  }
  \newgame
  \fenboard{8/8/2BBBB2/3B4/2B2b2/2B1bbb1/2Bbbbb1/8 w - - 0 1}

  \chessboard
  \caption{Key Squares and Diagonals for Bishops}
  \label{fig:key-bish}
\end{figure}

The applications of the model on Grandmaster games provide valuable insights into the dynamics and strategies employed by top-level chess players. By predicting the Centipawn evaluations $c(s)$ and winning probabilities $w(s)$ for specific subsets of states, we gain a deeper understanding of the advantages and disadvantages associated with different chess positions. These insights have several practical applications in chess analysis and gameplay evaluation.

The predictions generated by the model offer a quantitative measure of the advantage/disadvantage provided by the Knight and Bishop pieces in specific states. Heat maps depicting the predicted Centipawn evaluations $c(s)$ and winning probabilities $w(s)$ are presented for both White and Black knights and bishops. These visual representations provide a comprehensive overview of the relative strengths and weaknesses of these pieces in various positions.

By focusing on specific subsets of states, we can analyze the effectiveness of the Knight and Bishop pieces individually, as well as their contributions to the overall gameplay strategies employed by Grandmasters. This analysis aids in strategic decision-making, enabling players to assess the potential advantages or disadvantages associated with specific moves and piece configurations.

Furthermore, the expandability of the model allows for a comprehensive examination of the game across different states. By extending the analysis to include all pieces, we can uncover additional insights into the dynamics of the game and evaluate the effectiveness of various gameplay approaches. This broader perspective enhances our overall understanding of chess strategies and gameplay dynamics.

The predictions generated by the model can also be utilized for comparative analysis between different players or groups of players. By analyzing the Centipawn evaluations and winning probabilities associated with specific states, we can identify patterns and trends in the strategies employed by Grandmasters. This information can be leveraged to develop training materials and strategies for aspiring chess players, helping them improve their gameplay and decision-making abilities.

For example, in Figure \ref{fig:gm-wk-2}, where $w(s)$ represents the evaluation of the knight-square state, we can observe that the lowest values of $w(s)$ are found in the white corners of the chessboard, specifically squares a1 and h1. This observation aligns with the widely held belief that knights are generally considered as being in their worst positions when confined to the corners of the board.

The disadvantage of having a knight in the corner may stem from its limited mobility and restricted scope of influence. When placed in the corners, knights have fewer potential squares to reach and can easily become isolated from the central and more strategically significant areas of the board.

On the other hand, as the knights move closer to the opponent's side of the board, their positional value tends to increase. This is most likely due to the knights' ability to infiltrate enemy territory, potentially attacking key squares, pieces, or pawns.

The increasing value of knight-square states as the knights advance can be attributed to several factors. Firstly, the proximity to the opponent's pieces and pawns provides more targets for the knight's maneuvers and attacks. Secondly, knights positioned closer to the enemy's side can exert greater control over central squares and influence the dynamics of the game. This control can restrict the opponent's options and potentially create weaknesses in their position.

\begin{figure}[htb]
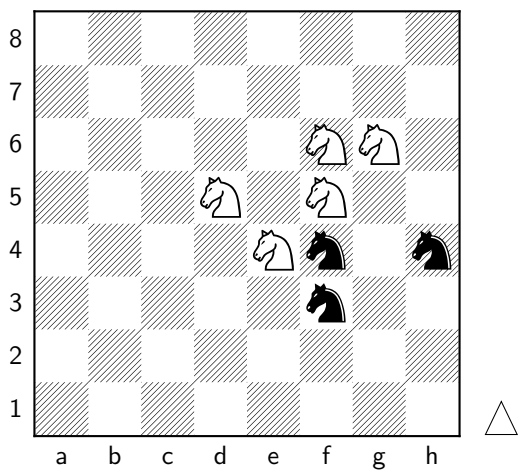

  \center
  \setchessboard{
    showmover=true,
    moverstyle=triangle,
    pgfstyle=color,
    pgfcolor=blue,
    markfield={},
    color=red,
    backfields={},
  }

  \newgame
  \fenboard{8/8/5NN1/3N1N2/4Nn1n/5n2/8/8 w - - 0 1}

  \chessboard
  \caption{Key Squares for Knights}
  \label{fig:key-knight-squares}
\end{figure}

Analyzing the values of knight-square states in different positions on the board, such as the corners and closer to the opponent's side, supports the claim that the placement of knights significantly affects their effectiveness. Understanding the strengths and weaknesses associated with different knight positions helps players make informed decisions about piece placement, strategic plans, and tactical considerations. Key squares for the knight to occupy are marked in Figure \ref{fig:key-knight-squares}.

The applications of our model on Grandmaster games provide valuable insights into the dynamics and strategies employed in high-level chess. The predictions of Centipawn evaluations and winning probabilities offer a quantitative measure of the advantages and disadvantages associated with specific chess positions, aiding in strategic decision-making and gameplay evaluation. The expandability of the model allows for a comprehensive analysis of the game across different states, facilitating a deeper understanding of chess strategies and enhancing the overall gameplay experience.

\subsection{Magnus Carlsen}
Our proposed model can be further applied to gain insights into the playing style and performance of specific players. In this section, we focus on the world-renowned chess player Magnus Carlsen (World Chess Champion, 2013–2023). By applying our model to the games played by Carlsen, we aim to uncover unique patterns and characteristics that contribute to his success and distinguish his gameplay from other Grandmasters.

\begin{figure}[H]
  \centering
  \begin{minipage}[b]{0.45\linewidth}
    \centering
    \includegraphics[width=\textwidth]{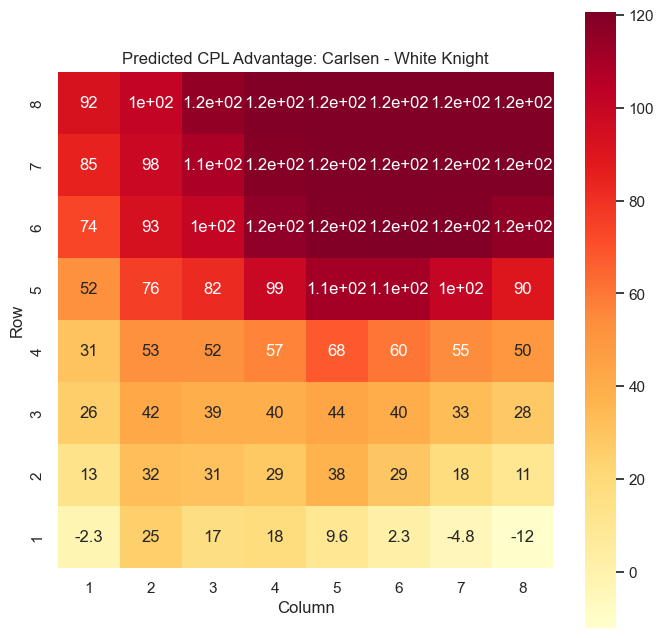}
    \caption{Predicted CPL Advantage $c(s)$ Offered by White Knight in Carlsen Games}
    \label{fig:c-wk-1}
  \end{minipage}
  \hspace{0.5cm}
  \begin{minipage}[b]{0.45\linewidth}
    \centering
    \includegraphics[width=\textwidth]{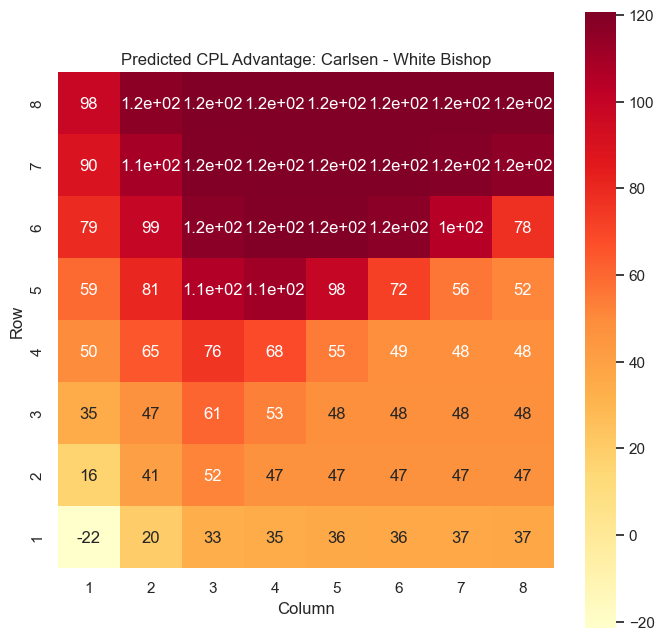}
    \caption{Predicted CPL Advantage $c(s)$ Offered by White Bishop in Carlsen Games}
    \label{fig:c-wb-1}
  \end{minipage}
\end{figure}

\begin{figure}[H]
  \centering
  \begin{minipage}[b]{0.45\linewidth}
    \centering
    \includegraphics[width=\textwidth]{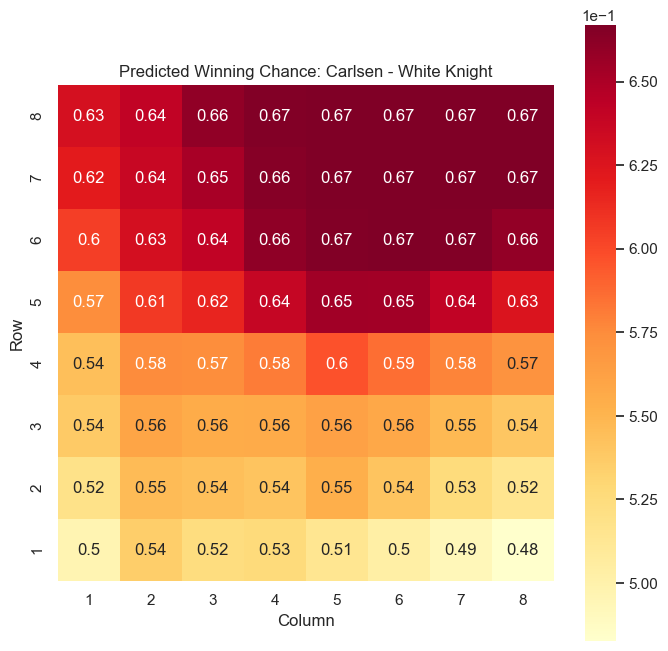}
    \caption{Predicted Winning Chance $w(s)$ Offered by White Knight in Carlsen Games}
    \label{fig:c-wk-2}
  \end{minipage}
  \hspace{0.5cm}
  \begin{minipage}[b]{0.45\linewidth}
    \centering
    \includegraphics[width=\textwidth]{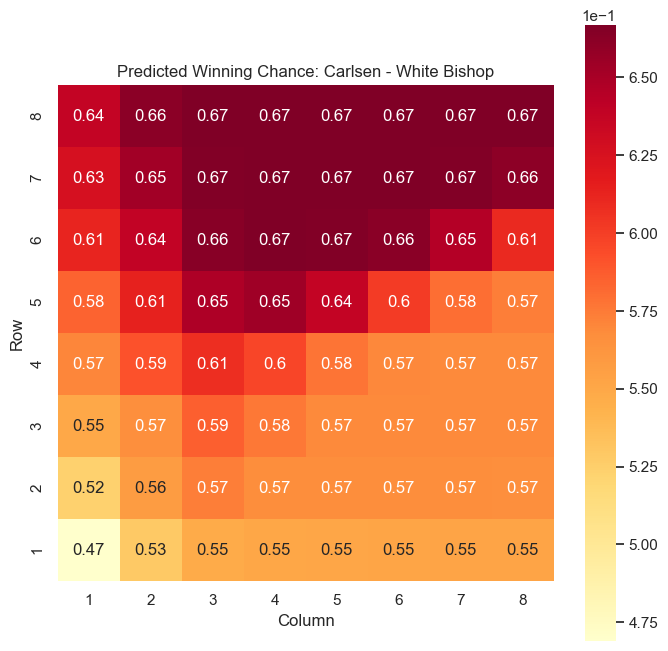}
    \caption{Predicted Winning Chance $w(s)$ Offered by White Bishop in Carlsen Games}
    \label{fig:c-wb-2}
  \end{minipage}
\end{figure}

Our proposed model is applied to a dataset consisting of 2000+ Carlsen games played in the last 5 years. Similar to the previous section, we begin by predicting the Centipawn evaluations $c(s)$ for states where Carlsen plays as the ``White" color and utilizes the ``Knight" or ``Bishop" piece. These predictions provide valuable insights into the relative advantages or disadvantages of Carlsen's chosen states, shedding light on his strategic decision-making process. The resulting heat maps, showcased in Figure \ref{fig:c-wk-1}, Figure \ref{fig:c-wb-1}, Figure \ref{fig:c-wk-2}, and Figure \ref{fig:c-wb-2}, offer a visual representation of the predicted Centipawn evaluations for Carlsen's specific subset of states.

Building upon this analysis, we further utilize the Centipawn evaluations $c(s)$ to derive the corresponding probabilities of winning $w(s)$ for Carlsen's selected states. The model-generated winning probabilities provide a clear representation of Carlsen's likelihood of winning the game given the occurrence of the specified state $s$.

By focusing on Carlsen's gameplay, we gain a deeper understanding of his preferred strategies and tendencies when employing the Knight piece as the ``White" color. This analysis allows us to assess the effectiveness of Carlsen's gameplay choices, providing insights into his decision-making process and potential areas of strength or improvement. Additionally, comparing Carlsen's results to the general dataset of Grandmaster games helps us evaluate his performance against the broader chess community.

\begin{figure}[H]
  \centering
  \begin{minipage}[b]{0.45\linewidth}
    \centering
    \includegraphics[width=\textwidth]{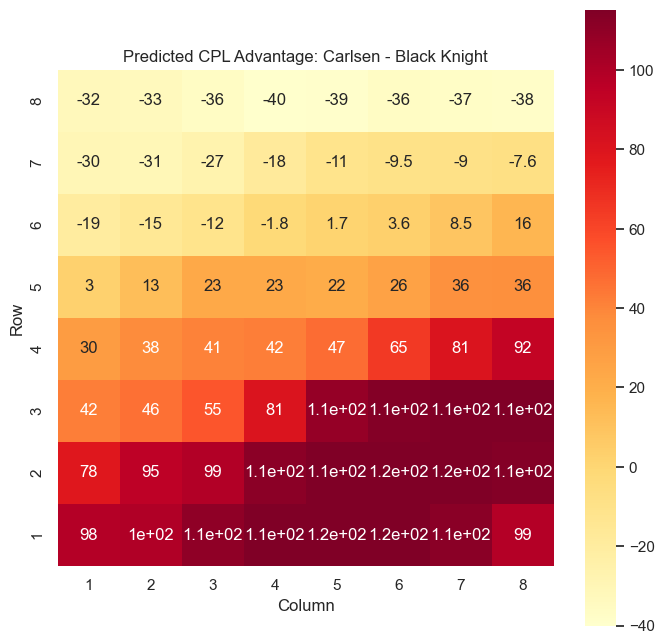}
    \caption{Predicted CPL Advantage $c(s)$ Offered by Black Knight in Carlsen Games}
    \label{fig:c-bk-1}
  \end{minipage}
  \hspace{0.5cm}
  \begin{minipage}[b]{0.45\linewidth}
    \centering
    \includegraphics[width=\textwidth]{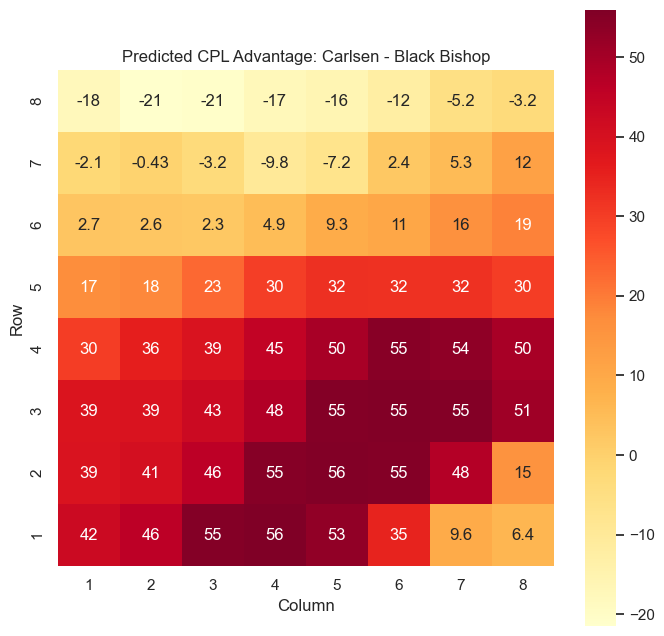}
    \caption{Predicted CPL Advantage $c(s)$ Offered by Black Bishop in Carlsen Games}
    \label{fig:c-bb-1}
  \end{minipage}
\end{figure}

\begin{figure}[H]
  \centering
  \begin{minipage}[b]{0.45\linewidth}
    \centering
    \includegraphics[width=\textwidth]{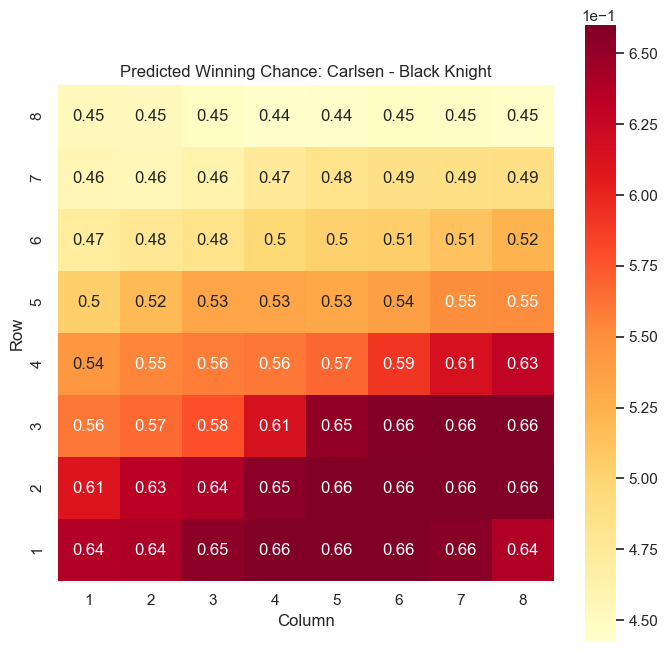}
    \caption{Predicted Winning Chance $w(s)$ Offered by Black Knight in Carlsen Games}
    \label{fig:c-bk-2}
  \end{minipage}
  \hspace{0.5cm}
  \begin{minipage}[b]{0.45\linewidth}
    \centering
    \includegraphics[width=\textwidth]{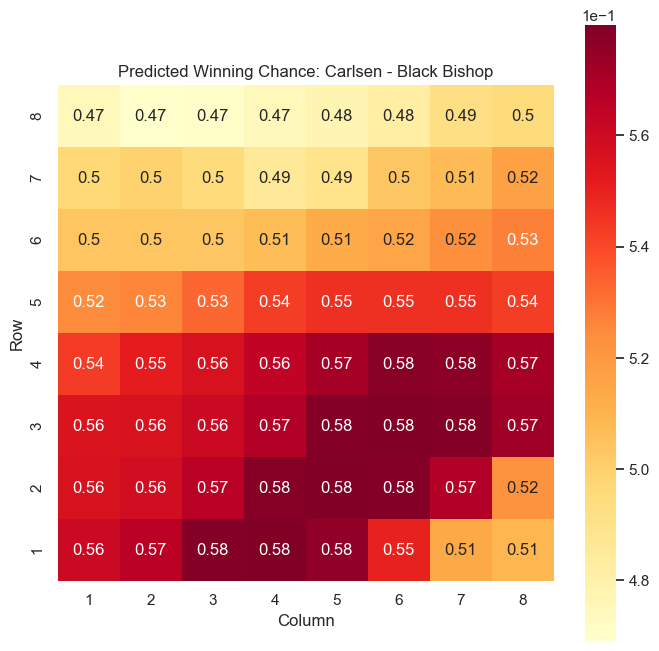}
    \caption{Predicted Winning Chance $w(s)$ Offered by Black Bishop in Carlsen Games}
    \label{fig:c-bb-2}
  \end{minipage}
\end{figure}

The model is then used to determine $c(s)$ and $w(s)$ for states ${(c, p, sq) \in s : c = \text{"Black"}, p = \text{{"Knight", "Bishop"}}}$, as can be seen in Figure \ref{fig:c-bk-1}, Figure \ref{fig:c-bb-1}, Figure \ref{fig:c-bb-2}, and Figure \ref{fig:c-bk-2}, respectively.

The applications of the model on Magnus Carlsen's games provide valuable insights into the dynamics and strategies employed by one of the world's top chess players. By predicting the Centipawn evaluations $c(s)$ and winning probabilities $w(s)$ for specific subsets of states, we can gain a deeper understanding of the advantages and disadvantages associated with different chess positions in Carlsen's games. These insights have numerous practical applications in chess analysis and gameplay evaluation.

The predictions generated by the model offer a quantitative measure of the advantage/disadvantage provided by the Knight and Bishop pieces in specific states encountered by Magnus Carlsen. Heat maps depicting the predicted Centipawn evaluations $c(s)$ and winning probabilities $w(s)$ are presented for both White and Black knights and bishops in Carlsen's games. These visual representations provide a comprehensive overview of the relative strengths and weaknesses of these pieces in various positions as encountered by Carlsen.

By focusing on specific subsets of states in Carlsen's games, we can analyze the effectiveness of the Knight and Bishop pieces individually, as well as their contributions to Carlsen's overall gameplay strategies. This analysis aids in strategic decision-making, enabling players to assess the potential advantages or disadvantages associated with specific moves and piece configurations based on Carlsen's approach.

Furthermore, the expandability of the model allows for a comprehensive examination of the game across different states in Carlsen's games. By extending the analysis to include all pieces, we can uncover additional insights into the dynamics of the game as played by Carlsen and evaluate the effectiveness of various gameplay approaches employed by him. This broader perspective enhances our overall understanding of Carlsen's strategies and gameplay dynamics.

The predictions generated by the model can also be utilized for comparative analysis between Magnus Carlsen and other players. By analyzing the Centipawn evaluations and winning probabilities associated with specific states in Carlsen's games, we can identify patterns and trends in his strategies. This information can be leveraged to develop training materials and strategies for aspiring chess players, helping them improve their gameplay and decision-making abilities while considering Carlsen's approach.

\begin{figure}[H]
  \centering
  \includegraphics[width=0.5\textwidth]{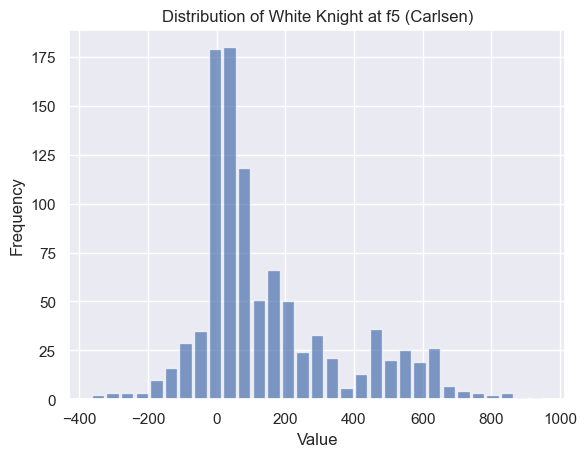}
  \caption{Histogram of $c(s)$ for White Knight on f5 for Carlsen}
  \label{fig:c-freq-f5}
\end{figure}

In Figure \ref{fig:c-wk-1}, we discover the solution to one of the questions raised in Section \ref{sec:intro}: the value of the white knight on f5. Figure \ref{fig:c-freq-f5} illustrates the distribution of $c(s)$ for the White Knight on f5 in Carlsen's games. It is evident that the $c(s)$ values for the White Knight exhibit a positive skew, indicating that this particular state $s$ is typically associated with favorable $c(s)$ values. Therefore, having a white knight positioned on f5 often confers an advantage.

By incorporating such insights into our analysis of Carlsen's games, we gain a more comprehensive understanding of the strengths, weaknesses, and strategic implications of the Knight and Bishop pieces as employed by Magnus Carlsen.

In sum, the applications of our model on Magnus Carlsen's games provide valuable insights into the dynamics and strategies employed by this world-class chess player. The predictions of Centipawn evaluations and winning probabilities offer a quantitative measure of the advantages and disadvantages associated with specific chess positions encountered by Carlsen, aiding in strategic decision-making and gameplay evaluation. The expandability of the model allows for a comprehensive analysis of Carlsen's games, facilitating a deeper understanding of his strategies and enhancing the overall gameplay experience.

\section{Pawn Valuation}\label{sec:pawns}

\vspace{0.1in}

\emph{No pawn exchanges, no file-opening, no attack---Aron Nimzowitsch}

\vspace{0.1in}

Our study is not complete until we apply the model to the mighty pawn. Our proposed model is applied to a comprehensive dataset comprising over 2000 Grandmaster games. The primary objective is to predict the probabilities of winning $w(s)$ and Centipawn evaluations $c(s)$ for a specific subset of states, namely those denoted by $\{ (c, p, sq) \in s : p \in \{\text{{Pawn}}\} \}$.

\begin{figure}[H]
  \centering
  \includegraphics[width=0.5\textwidth]{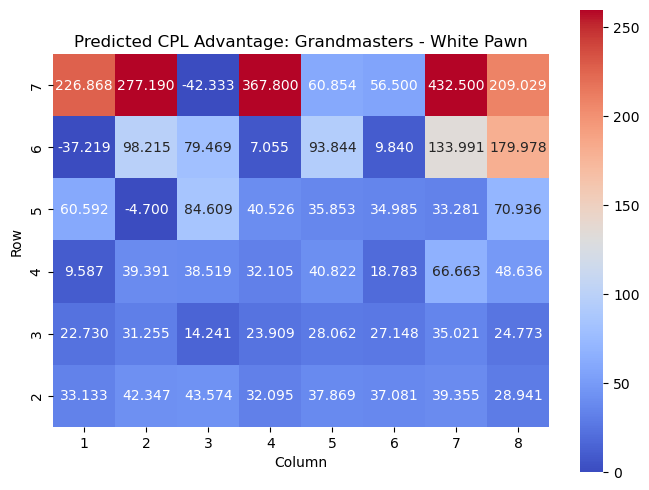}
  \caption{Predicted CPL Advantage $c(s)$ Offered by White Pawn in Grandmaster Games}
  \label{fig:pw-1}
\end{figure}

\begin{figure}[H]
  \centering
  \includegraphics[width=0.5\textwidth]{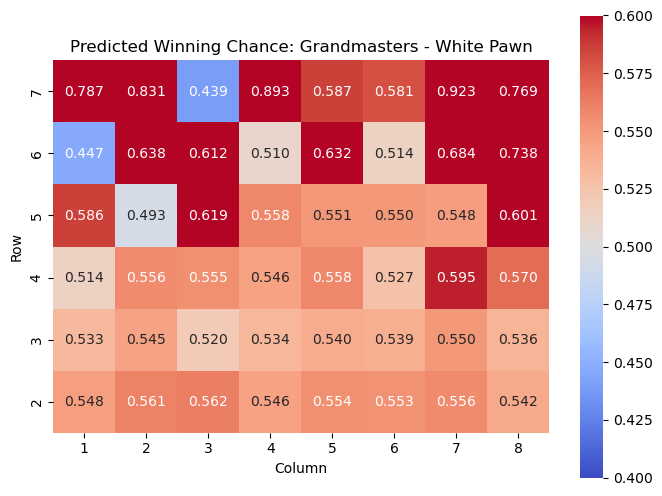}
  \caption{Predicted Winning Chance $w(s)$ Offered by White Pawn in Grandmaster Games}
  \label{fig:pw-2}
\end{figure}

The results of the model when applied to the White Pawn are shown in Figure \ref{fig:pw-1} and Figure \ref{fig:pw-2}.

We note a few chess maxims that are reflected in the model predictions.

\begin{enumerate}
  \item Pawns gain in value as they cross the 4th rank: This point highlights an important principle in chess, where advancing pawns beyond the 4th rank often leads to increased positional strength and potential threats. As pawns move forward, they gain control over more squares, restrict the opponent's piece mobility, and open up lines for their own pieces. Crossing the 4th rank is a significant milestone that can significantly impact the dynamics of the game.

  \item Pawns on the \textbf{h} and \textbf{a} files are very good on the 5th rank: This point emphasizes the strategic importance of pawns positioned on the h and a files when they reach the 5th rank. Pawns on these files can have a powerful influence on the game, particularly in the endgame. Placing pawns on the 5th rank provides support for the central pawns, helps control key central squares, and may facilitate piece activity and potential attacks on the opponent's position.

  \item Pawns on the 6th rank are deadly, especially when supported by a pawn on the 5th rank: This point highlights the strength of pawns on the 6th rank, which is just two steps away from promotion. Pawns advanced to this rank become highly dangerous, as they pose a direct threat to promote to a more powerful piece. When supported by a pawn on the 5th rank, these pawns can create a formidable pawn duo, exerting significant pressure on the opponent's position and potentially leading to advantageous tactical opportunities.

  \item Edge pawns tend to be weaker than central pawns: This point draws attention to the relative weakness of pawns placed on the edges of the board (such as the a and h files) compared to pawns in central positions. Edge pawns have fewer potential squares to advance or support other pieces, limiting their mobility and influence. In contrast, central pawns control more critical squares, contribute to a stronger pawn structure, and have a greater impact on the overall game dynamics.

\end{enumerate}

We also found chess ideas that are not considered to be common knowledge. For example, Kingside pawns are more dangerous when advanced than queenside pawns. We attempt to offer some anecdotal rationale for this: advancing pawns on the kingside (\textbf{g} and \textbf{h} files for White) can have a more immediate and aggressive impact compared to advancing pawns on the queenside (\textbf{a} and \textbf{b} files for White). Advanced kingside pawns can create open lines, potentially exposing the opponent's king to attacks or weakening their pawn structure. Understanding this distinction helps players assess the strategic implications of pawn advances on different sides of the board.

Important squares for the white pawn can also be seen by examining the highest Centipawn evaluation $c(s)$ values in each column. By analyzing the rows in the heatmap corresponding to the white pawns, we can identify squares that consistently have high Centipawn evaluations, indicating their significance for white pawns.

Starting from the top row (from White's perspective), the squares with the highest $c(s)$ values are $e4$, $h4$, $c5$, and $h6$. These squares represent critical positions for white pawns.

The square $e4$, located in the fourth row, is a well-known central square in chess. Occupying $e4$ with a white pawn can provide several advantages, such as controlling important central squares, supporting piece development, and establishing a strong pawn presence in the center.

Also in the fourth row, we find the square $h4$. Although it is on the edge of the board, it is an important square for white pawns. Placing a pawn on $h4$ can serve multiple purposes, including potentially supporting a kingside pawn storm, reinforcing control over the $g5$ square, or preparing to launch an attack on the opponent's position.

In the fifth row, we encounter the square $c5$. Occupying $c5$ with a white pawn can contribute to a solid pawn structure and provide control over central squares. It may also support piece mobility and influence the game's dynamics, particularly in the context of pawn breaks or central pawn exchanges.

Finally, in the sixth row, the square $h6$ stands out with the highest $c(s)$ value. Placing a pawn on $h6$ can have strategic implications, such as potentially supporting kingside attacks or acting as a defensive shield for the king.

By identifying these squares with high $c(s)$ values, we gain valuable insights into the strategic positioning of white pawns. These squares offer opportunities for central control, piece activity, attacking potential, and overall pawn structure. Understanding the significance of these squares helps players make informed decisions regarding pawn placement, pawn breaks, and strategic plans to maximize their advantage in the game.

\begin{figure}[H]
  \centering
  \includegraphics[width=0.5\textwidth]{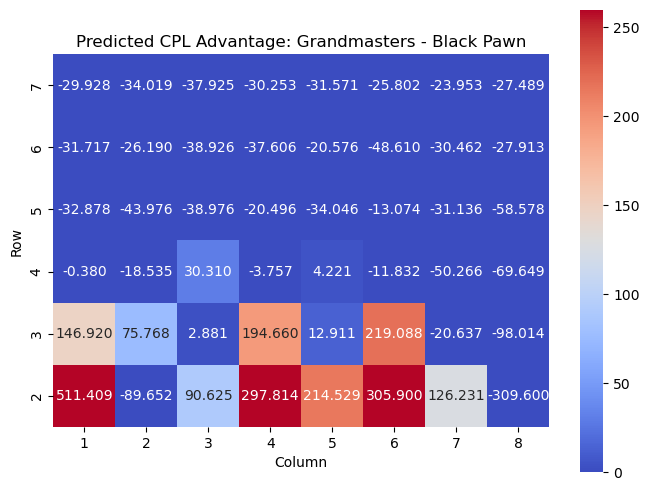}
  \caption{Predicted CPL Advantage $c(s)$ Offered by Black Pawn in Grandmaster Games}
  \label{fig:pb-1}
\end{figure}

\begin{figure}[H]
  \centering
  \includegraphics[width=0.5\textwidth]{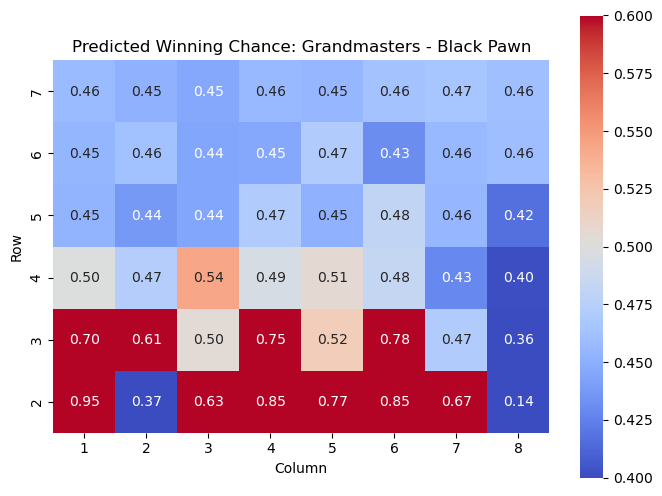}
  \caption{Predicted Winning Chance $w(s)$ Offered by Black Pawn in Grandmaster Games}
  \label{fig:pb-2}
\end{figure}

We next apply this model to the black pawns. The results are shown in Figure \ref{fig:pb-1} and Figure \ref{fig:pb-2}.

Similar conclusions can be drawn for the black pawns. By analyzing the highest Centipawn evaluation $c(s)$ values in each column for the black pawns, we can identify the key squares that consistently have high evaluations, signifying their significance for black pawns.

Just like for the white pawns, the rows in the heatmap corresponding to the black pawns reveal important squares. The squares with the highest $c(s)$ values for black pawns are $f5$, $d5$, $c4$, $d3$, and $f3$. These squares play a crucial role in determining the strength and strategic positioning of the black pawns.

The square $f5$, located in the fifth row, emerges as one of the critical squares for black pawns. Placing a pawn on $f5$ can provide black with control over central squares, potential support for piece development, and opportunities for counterplay.

The square $d5$ stands out with a high $c(s)$ value. Occupying $d5$ with a black pawn contributes to central control, potentially restricts white's pawn breaks, and provides a solid foundation for black's pawn structure.

In the fourth row, the square $c4$ is identified as an important square for black pawns. Occupying $c4$ can offer black strategic advantages, such as central control, potential support for piece activity, and the creation of tactical opportunities.

Furthermore, the square $d3$ in the third row holds significance for black pawns. Placing a pawn on $d3$ strengthens black's central presence, potentially restricts white's pawn advancements, and helps solidify black's position in the center.

Lastly, the square $f3$ in the third row also demonstrates a high $c(s)$ value. Occupying $f3$ with a black pawn can support kingside counterplay, potentially restrict white's piece mobility, and offer opportunities for tactical operations.

Analyzing these key squares for black pawns, namely $f5$, $d5$, $c4$, $d3$, and $f3$, provides valuable insights into the strategic considerations and potential strengths of the black pawn structure. Occupying and controlling these squares strategically enhances black's control of central areas, supports piece coordination, and enables counterplay against white's position.

By understanding the significance of these squares, players can make informed decisions regarding pawn placement, pawn breaks, and strategic plans to maximize their potential advantage and navigate the complexities of the game from the black perspective.

\section{Conclusion}\label{sec:conclusion}
In this paper, we presented a comprehensive methodology for evaluating chess positions and predicting the probabilities of winning $w(s)$ and Centipawn evaluations $c(s)$. Our approach utilized a combination of Centipawn evaluation, $Q$-learning, and Neural Networks to capture the complex dynamics of the game and facilitate informed decision-making.

We began by formalizing the theoretical functions used in $Q$-learning, such as the value function $V(s)$ and Centipawn evaluation $c(s)$. The value function represented the probability of winning the game given a specific state $s$, while the Centipawn evaluation measured the advantage in chess. We derived the win probability $w(s)$ from the Centipawn evaluation using a mathematical equation.

To address the sequential decision problem, we employed the dynamic programming technique of Q-learning, which involved breaking down the problem into smaller sub-problems and solving the Bellman equation. The $Q$-value matrix represented the probability of winning given a policy/move in a specific state, and we determined the optimal policy and value function using the $Q$-values.

To predict Centipawn evaluations $c(s)$, we designed a Neural Network architecture specifically tailored for chess positions. This model incorporated the $\tanh$ activation function to capture intricate patterns and relationships within the input data. By training the Neural Network on a meticulously crafted dataset, we could make accurate predictions of Centipawn evaluations for unseen states.

Our methodology expanded upon previous work by considering a comprehensive state representation that encompassed color, piece type, and square information. This allowed for a more nuanced analysis of the game dynamics and a deeper understanding of the factors influencing the outcome. We also showcased the applications of our model, focusing on specific subsets of states, such as the Knight and Bishop pieces, and visualizing the predicted probabilities of winning and Centipawn evaluations through heat maps.

Further research in this area could explore the dynamic nature of square values, taking into account positional changes and the interaction between different pieces. By refining and expanding our methodology, we can continue to deepen our understanding of the intricate dynamics of chess positions and contribute to advancements in the field of chess AI.

In conclusion, our methodology provides a robust framework for evaluating chess positions and making informed decisions during gameplay. By combining Centipawn evaluation, $Q$-learning, and Neural Networks, we achieved a comprehensive analysis of the game dynamics and enhanced our ability to assess strategic moves and guide decision-making. Our research contributes to the development of more sophisticated and intelligent Chess AI systems, paving the way for deeper insights into the intricacies of the game.

With our methodology, we strive to unravel the logical relations of chess and provide a comprehensive understanding of the game, empowering players and researchers alike to unlock new levels of strategic thinking and mastery.

\section{Author Contributions}
All authors have read and agreed to the published version of the manuscript.

\section{Funding}
This research received no external funding.
\section{Data Availability}
The data used in this study is available upon request from the corresponding author.
\section{Conflict of Interest}
The authors declare no conflict of interest.

\bibliographystyle{plainnat}
\bibliography{ref}
\end{document}